\documentclass[10pt,twocolumn,letterpaper]{article}

\usepackage{cvpr}    
\usepackage{times}
\usepackage{epsfig}
\usepackage{graphicx}
\usepackage{amsmath}
\usepackage{amssymb}

\usepackage{pifont}
\newcommand{\cmark}{\ding{51}}%
\newcommand{\xmark}{\ding{55}}%

\usepackage{color}
\usepackage{algorithm}
\usepackage{algpseudocode}
\usepackage{dsfont}
\usepackage{subcaption}
\usepackage{cite}
\usepackage{booktabs}
\usepackage{tabularx}
\usepackage{url}
\usepackage{soul}
\usepackage{balance}
\usepackage[normalem]{ulem}
\usepackage[accsupp]{axessibility}  

\usepackage{lscape} 
\usepackage{array,booktabs}

\usepackage[pagebackref,breaklinks,colorlinks,bookmarks=false]{hyperref}

\newcommand{\comment}[1]{}

\newcommand{\bx}[0]{\mathbf{x}}

\newcommand{\fig}[1]{Fig.~\ref{fig:#1}}

\newcommand{\algo}[1]{Alg.~\ref{alg:#1}}

\newcommand{\loss}{\mathcal{L}}

\definecolor{orange}{rgb}{1,0.5,0}
\definecolor{blue}{rgb}{0,0,0.6}
\definecolor{green}{rgb}{0,0.6,0}
\definecolor{purple}{rgb}{0.5,0,0.5}

\definecolor{color1}{RGB}{0,199,1}
\definecolor{color2}{RGB}{224,43,28}

\newcommand{\btheta}{\boldsymbol{\theta}}

\newcolumntype{P}[1]{>{\centering\arraybackslash}p{#1}}

\renewcommand{\paragraph}[1]{\vspace{0.5em}\noindent\textbf{#1}}

\def\cvprPaperID{3210} 
\def\confName{CVPR}
\def\confYear{2022}

\begin{document}

\title{Collaborative Learning for Hand and Object Reconstruction with Attention-guided Graph Convolution}

\author{Tze Ho Elden Tse$^1$ \qquad Kwang In Kim$^2$ \qquad Ale\u{s} Leonardis$^1$ \qquad Hyung Jin Chang$^1$\\
$^1$University of Birmingham \qquad $^2$UNIST\\
{\tt\small txt994@student.bham.ac.uk, kimki@unist.ac.kr, \{a.leonardis, h.j.chang\}@bham.ac.uk}
}

\maketitle

\begin{abstract}
Estimating the pose and shape of hands and objects under interaction finds numerous applications including augmented and virtual reality.
Existing approaches for hand and object reconstruction require explicitly defined physical constraints and known objects, which limits its application domains.
Our algorithm is agnostic to object models, and it learns the physical rules governing hand-object interaction. This requires automatically inferring the shapes and physical interaction of hands and (potentially unknown) objects.
We seek to approach this challenging problem
by proposing a collaborative learning strategy where 
two-branches of deep networks 
are learning from each other. 
Specifically, we transfer hand mesh information to the object branch and vice versa for the hand branch. 
The resulting optimisation (training) problem can be unstable, and we address this via 
two strategies: 
(i) attention-guided graph convolution which helps identify and focus on mutual occlusion and (ii) unsupervised associative loss which facilitates the transfer of information between the branches.
Experiments using four widely-used benchmarks
show that our framework achieves beyond state-of-the-art
accuracy in 3D pose estimation, as well as recovers
dense 3D hand and object shapes. Each technical component above contributes meaningfully in the ablation study. 
\end{abstract}

\vspace{-0.1cm}
\section{Introduction}
\begin{figure}
\centering
\includegraphics[width=0.9\linewidth]{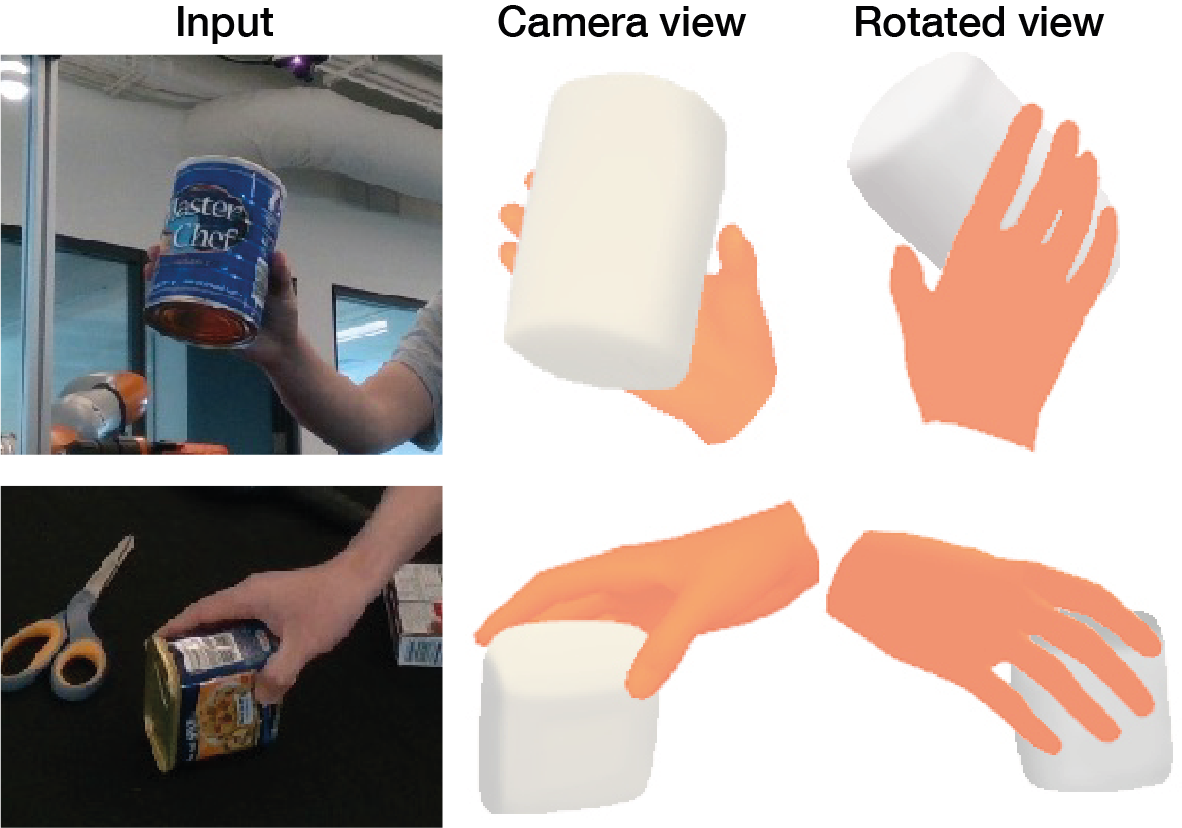}
\vspace{-0.2cm}
\caption{We propose a collaborative learning framework which allows sharing of mesh information across hand and object branches iteratively. Our model jointly reconstructs hand and object meshes from a monocular RGB image.
}
\vspace{-0.5cm}
\label{fig:teaser}
\end{figure}

Understanding human hand and object interaction is fundamental for meaningful interpretation of human action and behaviour \cite{tekin2019h+,wang2020learning}. With the advent of deep learning and RGB-D sensors, pose estimation of isolated hands has made significant progress, \eg, depth-based \cite{wan2017crossing,yuan2017bighand2,wu2017hand,cai2018weakly,yuan2018depth} and RGB-based \cite{simon2017hand,zimmermann2017learning,mueller2018ganerated,spurr2018cross,yang2020seqhand} methods. 
However, despite a strong link to real applications such as
augmented and virtual reality
\cite{han2020megatrack,wang2020rgb2hands,mueller2019real}, joint reconstruction of hand and object \cite{hasson2019learning,hasson2020leveraging} has received relatively less attention. 
In this paper, we focus on the problem of hand and object reconstruction from a single RGB image (see \fig{teaser}). 

Joint hand and object pose estimation is a challenging problem. First, while self-occlusion in hand is a well-known problem~\cite{ye2018occlusion,rangesh2016hidden}, when interacting with objects, hands (and objects) exhibit even greater occlusion from almost any point of view mutually \cite{nakamura2017complexities}. 
Secondly, first-person-view (\eg, \emph{FHB} \cite{FirstPersonAction_CVPR2018} dataset) often exhibits 
large degree of erratic camera motion. Recent works \cite{tekin2019h+,doosti2020hope,liu2021semi} have been able to tackle some major challenges in joint hand-object pose estimations in colour input.
However, in the absence of physical constraints, and with sparse keypoint detection, they often lead to erroneous pose estimation or mesh reconstructions (\eg hands penetrating objects).

To fundamentally 
understand hand-object interactions, it is essential to fully recover 3D information, and accordingly, there has been significant improvements towards hand mesh estimations from single RGB image \cite{ge20193d,baek2019pushing,boukhayma20193d,zhang2019end,zimmermann2019freihand,kulon2020weakly,zhou2020monocular,baek2020weakly,choi2020pose2mesh,moon2020deephandmesh}. 
Hasson~\etal~\cite{hasson2019learning} further proposed attraction and repulsion loss terms to generate physically plausible reconstructions. Recent optimisation-based approaches \cite{cao2020reconstructing,hasson20_handobjectconsist} that rely on these contact terms are limited to scenarios where hand and object are already in contact. However, the ability to reason pre-grasp stages are equally important as it allows robots to infer human intents \cite{meltzoff1995understanding} and learn manipulation skills from humans \cite{mandikal2020dexterous}. Therefore, we propose a 
strategy that is not restricted by these contact terms and is able to learn the context of actual as well as near physical contact.


Our novel collaborative learning framework allows hand and object branches to boost each other in a progressive and iterative fashion. There are two motivations for this 
strategy: 
1) estimating the pose of interacting hands and objects is a highly-correlated task and 2) mutual occlusions can be tackled by simultaneously sharing mesh information. This is supported by the fact that the image encoder struggles to extract useful features under mutual occlusion, and therefore capturing object mesh information would compensate this limitation for hand reconstruction (the same in object branch). Previous attempts in this context share information across branches via simple branch stacking \cite{yang2020collaborative} where communication bottleneck exists: We empirically observed that performance gain across network inference iterations are limited in this approach. We explicitly address this by a new unsupervised associative loss facilitating the information transfer.
Further, to address frequently occurring occlusions in hand-object interaction scenarios, we propose an attention-guided graph convolution that 
can be trained in an unsupervised manner. Our graph convolution demonstrates the ability to improve mesh quality as well as correct hand and object poses.

Our contributions are the following: 
\begin{enumerate}
    \item We propose an end-to-end trainable collaborative learning strategy for hand-object reconstruction from a single RGB image. \vspace{-0.7em}
    \item We design an attention-guided graph convolution to capture mesh information dynamically.\vspace{-0.7em}
    \item We introduce an unsupervised training strategy for effective feature transfer between hand-object branches.\vspace{-0.7em}
    \item We demonstrate that our model achieves highly physically plausible results without contact terms.
\end{enumerate}
We evaluate our method on four hand-object datasets \ie \emph{FHB} \cite{FirstPersonAction_CVPR2018}, \emph{ObMan} \cite{hasson2019learning}, \emph{HO-3D} \cite{hampali2020honnotate} and \emph{DexYCB} \cite{chao2021dexycb} and demonstrate that our method significantly outperform state-of-the-art approaches.

\begin{figure*}[ht]
\centering
\includegraphics[width=0.99\linewidth]{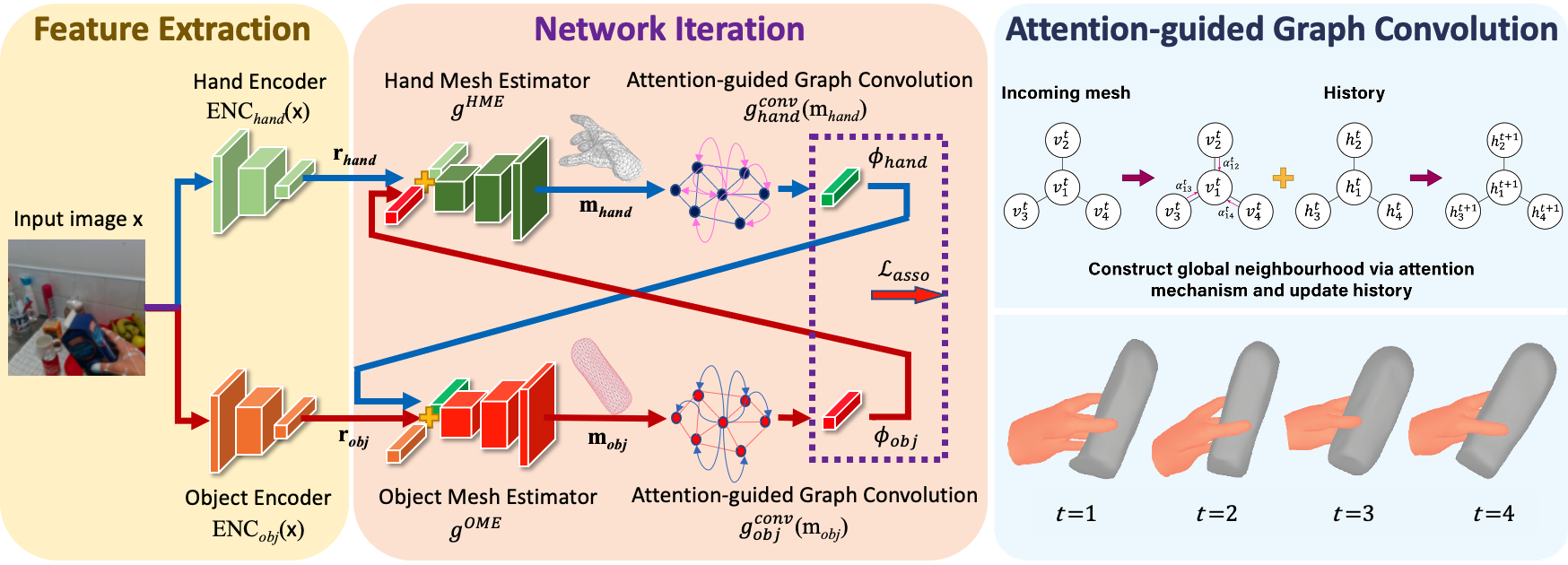}
\vspace{-0.2cm}
\caption{A schematic illustration of our framework. It takes an input image $\textbf{x}$, which goes through two separate ResNet-18 \cite{he2016deep} encoders, $\text{ENC}_{hand}(\mathbf{x})$ and $\text{ENC}_{obj}(\mathbf{x})$ to produce hand and object features, $\mathbf{r}_{hand}$ and $\mathbf{r}_{obj}$, respectively. Hand mesh estimator $g$\textsuperscript{\textit{HME}} takes $\mathbf{r}_{hand}$ and output hand mesh $\mathbf{m}_{hand}$ which is then pass to graph convolution module $g_{hand}^{conv}$ and output $\phi_{hand}$. Object mesh estimator takes both $\mathbf{r}_{obj}$ and $\phi_{hand}$ to output object mesh $\mathbf{m}_{obj}$. Similarly, graph convolution module $g_{obj}^{conv}$ takes object mesh $\mathbf{m}_{obj}$ and output $\phi_{obj}$ which is then combine with hand features $\mathbf{r}_{hand}$ and goes into the hand mesh estimator $g$\textsuperscript{\textit{HME}}. An unsupervised associative loss is used to supervise the feature transfer process under network iterations, \ie $\phi_{hand}$ and $\phi_{obj}$. We have included an example on the bottom right corner which demonstrates the effect of our attention-guided graph convolution for iteration $t$.
}
\vspace{-0.3cm}
\label{fig:framework}
\end{figure*} 
\vspace{-0.2cm}
\section{Related works}

Our work tackles the problem of hand and object reconstruction from a single RGB image. 
We first review the literature on \emph{Hand-Object Reconstruction}. Then, we focus on the line of work that leverages \emph{Graph Convolutional Neural Networks} on hand-related tasks. Finally, we provide a brief review on \emph{Collaborative Learning} despite its weak link in the literature. 

\vspace{-0.1cm}
\paragraph{Hand-object reconstruction.} 
Joint reconstruction of hands and objects has been receiving increasing attention \cite{hasson2019learning,hasson2020leveraging,cao2020reconstructing,hasson20_handobjectconsist}. Hasson \etal~\cite{hasson2019learning} leverages a differentiable MANO network layer enabling end-to-end learning of hand shape estimation and incorporates contact losses which encourages contact surfaces and penalises penetrations between hand and object. Hasson \etal~\cite{hasson2020leveraging} assumes known object models and leverages photometric consistency as self-supervision on the unannotated intermediate frames to improve hand and object reconstructions. Karunratanakul \etal~\cite{karunratanakul2020grasping} proposes an implicit representation for hand in the form of sign distance fields. 
Recent works mostly adopt optimisation-based procedures to jointly fit hand-object meshes \cite{cao2020reconstructing,yang2021cpf,hasson20_handobjectconsist}. In this paper, we propose a learning-based strategy where immediate features are shared across hand-object branches and are able to produce physically plausible interactions without any contact terms.

\vspace{-0.1cm}
\paragraph{Graph convolution-based methods.}
As skeleton can be represented in a form of graph, graph convolution naturally attracts much attention in hand pose estimation. Graph convolutional neural networks (GCN) can be split into spectral- \cite{bruna2013spectral,defferrard2016convolutional,kipf2016semi} and spatial-based methods \cite{gilmer2017neural,xu2018powerful,monti2017geometric}.
For spectral-based application, \cite{ge20193d,choi2020pose2mesh} adopt the Chebyshev spectral graph convolution \cite{defferrard2016convolutional} to compute hand mesh.
Cai \etal~\cite{cai2019exploiting} leverages GCN \cite{kipf2016semi} and apply on the sequence of skeletons as a spatial-temporal graph to exploit the spatial and temporal consistencies for pose estimation. Doosti \etal~\cite{doosti2020hope} proposes a lightweight graph convolutional network which jointly estimates hand and object poses. Kulon \etal~\cite{kulon2020weakly} proposes spiral filters to recover hand mesh directly from autoencoder. They demonstrate that spatial mesh convolutions outperform spectral methods and SMPL-based models \cite{loper2015smpl,romero2017embodied} for hand reconstruction. In contrast, our proposed attention-guided graph convolution is able to take dynamic graph input and does not assume a fixed neighbourhood for feature aggregation.  

\vspace{-0.1cm}
\paragraph{Collaborative learning.}
There has been a lot of literatures concerning learning multiple tasks simultaneously. They span across the spectrum of multi-task learning \cite{baxter1997bayesian,baxter2000model,caruana1993multitask}, domain adaptation \cite{mansour2009domain,mansour2008domain}, distributed learning \cite{balcan2012distributed,dekel2011optimal,wang2016distributed} and collaborative learning\cite{blum2017collaborative,jiang2017collaborative,song2018collaborative,nguyen2018improved}. Collaborative learning refers to making learning more efficient through sharing of information. 
Blum \etal~\cite{blum2017collaborative} proposes a collaborative PAC (\textit{probably approximately correct}) learning model which was built upon Valiant \etal~\cite{valiant1984theory} and \cite{nguyen2018improved,chen2018tight} are the follow-up works. 
Song \etal~\cite{song2018collaborative} introduces one form of collaborative learning framework in which multiple classifier heads of the same network are simultaneously trained on the same training data to improve generalisation and robustness without extra inference cost. There are two major mechanisms under his framework: 1) Same training datasets for multiple views from different classifiers improves generalisation and 2) Intermediate-level representation sharing. Yang \etal~\cite{yang2020collaborative} exploits joint-aware features for gesture recognition and 3D hand pose estimation. Their mechanism focuses on intermediate-level representation sharing iteratively across multiple tasks. In this paper, we improve on \cite{yang2020collaborative} with an attention-guided graph convolution and an unsupervised associative loss to guide the intermediate-level representation sharing process. Also, our proposed graph convolution is based on a multi-head attention mechanism which possesses the spirit of \cite{song2018collaborative} to improve generalisation with multiple views on the same dataset.
\section{Collaborative estimation of hand and object meshes}
Our training pipeline, as shown in \fig{framework}, takes an input RGB image $\bx \in \mathbb{R}^{256\times256}$ and involves 4 steps for one iteration: 
1) Reconstruct hand mesh using the parametric MANO model \cite{romero2017embodied}; 2) Extract hand features from hand mesh guided by our associative loss; 3) Reconstruct object mesh by fusing object encoder features and extracted hand features from the previous step; and 4) Extract object features from object mesh.
Our architecture is split into hand and object branches. Each branch has a ResNet-18 \cite{he2016deep} encoder pre-trained on ImageNet \cite{russakovsky2015imagenet}: $\text{ENC}_{hand}(\bx)$ and $\text{ENC}_{obj}(\bx)$.

The key motivation for our approach is to leverage the implicit hand-object relationship: We target the problem of mutual occlusion in hand-object interactions by simultaneously sharing 3D reconstructions under our collaborative learning framework. However, na\"ively connecting network branches tended to accumulated errors, leading to highly unstable training. 
Therefore, we propose an attention-guided graph convolution to capture 3D reconstructions dynamically. In addition, by following the notion that hand shape deforms according to object shape, we propose an unsupervised associative loss to improve the feature transfer process from hand to object, and vice versa. Our networks are trained in an end-to-end manner. \algo{optim} summarises the training process.

\subsection{Hand mesh estimator $g$\textsuperscript{\textit{HME}}} \label{sec:HME}

We adopted the differential MANO \cite{romero2017embodied} model from \cite{hasson2019learning}. 
It maps pose ($\btheta\in\mathbb{R}^{51}$) and shape ($\boldsymbol{\beta}\in\mathbb{R}^{10}$) parameters to a mesh with $N=778$ vertices.  Pose parameters ($\btheta$) consists of $45$ DoF (\ie $3$ DoF for each of the $15$ finger joints) plus $6$ DoF for rotation and translation of the wrist joint. 
Shape parameters ($\boldsymbol{\beta}$) are fixed for a given person. A kinematic tree is formed with the $15$ joints and the wrist joint as the first parent node. Joint locations can be obtained using the kinematic tree with global rotation based on $\btheta$.

Given the $512$-dimensional hand feature vector $\mathbf{r}_{hand}$, we use a fully connected layer to regress $\btheta$ and $\boldsymbol{\beta}$. The original MANO model uses $6$-dimensional PCA (principal component analysis) subspace of $\btheta$
for computational efficiency. However, we empirically observed that full $45$-dimensional pose space better captures 
a variety of hand poses especially over sequential datasets. A hand mesh can be defined as $\mathbf{m}_{hand} = (\mathbf{v}_{hand},\mathbf{f}_{hand})$, where $\mathbf{v}_{hand}\in\mathbb{R}^{778\times3}$ refers to a set of vertices in the mesh and $\mathbf{f}_{hand}\in\mathbb{R}^{1538\times3}$ refers to a close set of edges (\ie a triangle face has $3$ edges). The mesh faces $\mathbf{f}_{hand}$ is provided by MANO \cite{romero2017embodied}.

\paragraph{Hand reconstruction loss $\loss_{hand}$.}
We directly optimise root-relative 3D positions by minimising their L2 distance to the corresponding ground-truth vertex positions $\mathbf{v}_{hand}^{*}$:
\vspace{-0.1cm}
\begin{align}
\label{e:handreconloss}
    \loss_{V} (\mathbf{v}_{hand})= \left\|\mathbf{v}_{hand}-\mathbf{v}_{hand}^{*}\right\|_{2}^{2}.
\end{align}

When ground truth vertex positions are not available, we supervise on 3D joint locations $\mathbf{J}\in\mathbb{R}^{n\times3}$ where $n$ refers to the number of joints. The 3D joint loss is defined as:
\vspace{-0.1cm}
\begin{align}
\label{e:lhandj}
    \loss_{J}(\mathbf{J}) = \left\|\mathbf{J}-\mathbf{J^{*}}\right\|_{2}^{2},
\end{align}
where $\mathbf{J^{*}}$ refers to ground truth joint positions. The resulting loss is defined as: $\loss_{hand} = \loss_{V} + \loss_{J}$.

We do not adopt hand shape regularisation as in \cite{hasson2019learning} as we empirically observed that our iterative process already prevents extreme mesh deformation.

\subsection{Object mesh estimator $g$\textsuperscript{\textit{OME}}} \label{sec:OME}
Given the $512$-dimensional object feature vector $\mathbf{r}_{obj}$, we adopt AtlasNet \cite{groueix2018papier} from \cite{hasson2019learning} to estimate object mesh $\mathbf{m}_{obj} = (\mathbf{v}_{obj},\mathbf{f}_{obj})$, \ie
$\mathbf{v}_{obj}\in\mathbb{R}^{642\times3}$ refers to object vertices and $\mathbf{f}_{obj}\in\mathbb{R}^{1280\times3}$ refers to object mesh faces.

\paragraph{Object reconstruction loss $\loss_{obj}$.}
As object mesh is reconstructed in the camera coordinate frame, it can be directly optimised by minimising the Chamfer distance as in \cite{groueix2018papier}. The resulting loss is defined as:
\vspace{-0.1cm}

\begin{align}
    \loss_{obj}(\mathbf{v}_{obj}) = &\frac{1}{2}\Big(\sum_{x \in \mathbf{v}_{obj}}d_{\mathbf{v}_{obj}^{*}}(x) + \sum_{y \in \mathbf{v}_{obj}^{*}}d_{\mathbf{v}_{obj}}(y)\Big), 
\end{align}
where $\mathbf{v}_{obj}^{*}$ refers to the points uniformly sampled on the surface of the ground truth object, $d_{\mathbf{v}_{obj}^{*}}(x)=\min_{y \in \mathbf{v}_{obj}^{*}}\left\|x-y\right\|_{2}^{2}$, and $d_{\mathbf{v}_{obj}}(y)=\min_{x \in \mathbf{v}_{obj}}\left\|x-y\right\|_{2}^{2}$.

\subsection{Attention-guided graph convolution $g^{conv}$} \label{sec:MC}

\paragraph{Preliminary.} We propose to use the message passing scheme \cite{gilmer2017neural} in graph convolution to capture mesh information and transfer to the opposite branch.
By denoting vertex feature $\mathbf{v}_{i}^{(k)}\in\mathbb{R}^{F}$ of vertex $i$ in layer $k$, the first step of such message passing scheme can be described as:
\vspace{-0.1cm}
\begin{align} \label{eq:aggregate}
    \mathbf{msg}_{i}^{k} = \text{AGGREGATE}^{(k)}\big(\big\{\mathbf{v}_{u}^{(k-1)}, u\in\mathcal{N}(i)\big\}\big),
\end{align}
where message $\mathbf{msg}_{i}^{k}$ is formed by aggregating neighbourhood $\mathcal{N}(i)$ around vertex $i$ from previous layer $(k-1)$.
The second step updates vertex feature with this new message:
\vspace{-0.1cm}
\begin{align} \label{eq:update}
    \mathbf{v}_{i}^{k} = \text{UPDATE}^{(k)}\big(\mathbf{v}_{i}^{(k-1)},\mathbf{msg}_{i}^{k}\big).
\end{align}
The choice for neighbourhood $\mathcal{N}(i)$, aggregating function $\text{AGGREGATE}^{(k)}$ and update function $\text{UPDATE}^{(k)}$ are crucial.  
There has been a variety of functions proposed in the literature \cite{defferrard2016convolutional,kipf2016semi,gilmer2017neural,xu2018powerful}. In this work, we propose to leverage attention mechanism to construct aggregating neighbourhood and a history term for updating node features.

\paragraph{Objective.} By defining $P$ to be the number of iterations per forward pass, the input is a sequence of meshes $(\mathbf{m}_{\theta}^{1}, \mathbf{m}_{\theta}^{2},...)$ where $\mathbf{m}_{\theta}^{t} = (\mathbf{v}_{\theta}^{t},\mathbf{f}_{\theta}^{t})$ for $t \in [1,\dots,P]$ is defined by vertices $\mathbf{v}_{\theta}^{t}$ and faces $\mathbf{f}_{\theta}^{t}$ for either branch $\theta \in \{hand,obj\}$. 
The objective is to estimate feature offset $\Delta_{hand}^{t}$ from the hand branch for object reconstruction, and vice versa:
\vspace{-0.1cm}
\begin{align}
    \mathbf{r}_{obj}^{t+1} = \mathbf{r}_{obj}^{t} + \Delta_{hand}^{t}.
\end{align}

\paragraph{Attention-guided graph convolution.}
As the above sequential task involves dynamically evolving graphs, static graph convolution would not be suitable because the weights are only being updated after $P$ iterations. Therefore, a solution should maintain the history of operations. Furthermore, our experiments confirm that static graph convolutions that assumes fixed neighbourhood do not benefit from increasing iterations $P$ (see Table \ref{table:ablation_P}).

By assuming input mesh vertices $\mathbf{v}_{\theta}$ is an un-ordered set, we propose to dynamically construct neighbourhoods $\mathcal{N}(i)$ using attention mechanism \cite{bahdanau2014neural,gehring2016convolutional}. Attention coefficient $\alpha_{ij} \in [0,1]$ is defined as the importance of vertex $j$’s features to vertex $i$ \cite{velickovic2018graph}. 
Node $j$ is included in the neighbourhood $\mathcal{N}(i)$ of $i$ when $\alpha_{ij}$ is larger than a threshold, \ie $0.5$.
Finally, our proposed graph convolution layer at iteration $t$ can be defined by rewriting Eqs.~(\ref{eq:aggregate}--\ref{eq:update}) as:
\vspace{-0.1cm}
\begin{align} \label{eq:cgc_agg}
    \alpha_{ij}^{t} = \frac{\exp\Big({\text{LeakyReLU}(\mathbf{a}^{\top}[\mathbf{W}\mathbf{v}_{i}^{t}\Vert\mathbf{W}\mathbf{v}_{j}^{t}])}\Big)}{\sum_{k\in\mathbf{v}^{t}}\exp\Big({\text{LeakyReLU}(\mathbf{a}^{\top}[\mathbf{W}\mathbf{v}_{i}^{t}\Vert\mathbf{W}\mathbf{v}_{k}^{t}])}\Big)}
\end{align}
where attention coefficient $\alpha_{ij}^{t}$ is computed using incoming vertices $\mathbf{v}^{t}=\{\mathbf{v}_{1}^{t},...,\mathbf{v}_{N}^{t}\}$ with $N$ being the maximum mesh vertices and learnable weights $\mathbf{a}\in\mathbb{R}^{2F}$ and $\mathbf{W}\in\mathbb{R}^{F \times 3}$. Note that $F$ is a hyperparameter and $\Vert$ is concatenation operation. We then update history $\mathbf{h}_{i}^{t}$ of vertex $i$:
\vspace{-0.1cm}
\begin{align} \label{cgc_update}
    \mathbf{h}_{i}^{t+1} = \text{LayerNorm} \bigg(\frac{1}{K}\sum_{k=1}^{K}\sum_{j\in \mathcal{N}^{t}(i)} \alpha_{ij}^{t,k} \mathbf{v}_{j}^{t}+ \mathbf{h}_{i}^{t} \bigg),
\end{align}
where $\mathcal{N}^{t}(i)$ is the aggregating neighbourhood around vertex $i$ at $t$, history $\mathbf{h}^{t}=\{\mathbf{h}_{1}^{t},...,\mathbf{h}_{N}^{t}\}$ and it is initialised as $\mathbf{0}$. Similar to \cite{velickovic2018graph,vaswani2017attention}, we find multi-head attention $\alpha_{ij}^{k}$ to be beneficial and apply layer normalisation \cite{ba2016layer} to stabilise and enable faster training. We use residual connection \cite{he2016deep} to track the history sequence and prevent performance drop on increasing iterations. In the final step, we use a fully connected layer to resize to the same size as image features $\mathbf{r}_{\theta}(\mathbf{x})$, namely $\phi_{\theta}$.

\paragraph{Discussions.} Our proposed graph convolution is reminiscent to GAT \cite{velickovic2018graph} and any $k$-nearest neighbours ($k$-NN) based dynamic graph convolutions like EdgeConv \cite{wang2019dynamic}. 
However, our approach differentiates from those because firstly, we do not assume static graph inputs.
Secondly, we differentiate from GAT \cite{velickovic2018graph} by how we leverage attention mechanism - they aggregate on fixed and local neighbourhood whereas we take this further by dynamically constructing global neighbourhood using attention mechanism. In addition, as the incoming mesh are 3D positions, $k$-NN like approaches suffer from local neighbourhood aggregation and high $k$-NN computational cost at each iterations. In short, our proposed method is able to capture long-range dependencies from dynamic graph in a single layer.
In Table~\ref{table:ablation_P}, we experiment with two common graph convolution operators (GCN \cite{kipf2016semi} and spiral mesh convolution \cite{gong2019spiralnet++,kulon2020weakly}) and demonstrate superior performance of our proposed attention-guided graph convolution.

\subsection{Associative supervision} \label{sec:AS}
Due to mutual occlusion in hand-object scenarios \cite{nakamura2017complexities}, it is challenging for the image encoder to capture useful information for mesh reconstruction. Instead, here we rely on the fact that hand pose changes with respect to different objects. For example, we hold cups differently depending on whether it has handle or not. We hypothesise that object branch benefits from hand mesh information (and vice versa for hand branch) and assume that good feature transfer in collaborative learning occurs when these features are 
highly similar within the same object class and distinctive across all other object classes. However, in practice, such object class information is not available. Hence, we propose an unsupervised loss to facilitate effective feature transfer.

Given $\phi_{\theta} = \{\phi_{\theta}^{1},...,\phi_{\theta}^{B}\}$ with $B$ being the input batch size, we update the image features by simple addition. In the following, we describe an unsupervised loss for $\phi_{\theta}$.

\paragraph{Associative loss $\loss_{asso}$.}
Our approach is inspired by \cite{haeusser2017learning} which was originally designed for semi-supervised learning. We imagine a walker going along $\Phi_{i} = [\phi_{hand}^{i};\phi_{obj}^{i}]$ where $i \in \{1,\ldots,B\}$.
As each $\Phi_{i}$ comes in pair with the same object class, we define a correct walk if transition is under the same object class. We 
define similarity between two embeddings as:
\vspace{-0.1cm}
\begin{align}
    M_{ij} = \Phi_{i}^\top\Phi_{j}, \;\; 1\leq i,j \leq B.
\end{align}
A single transition based on embeddings similarity is defined as:
\vspace{-0.1cm}
\begin{align}
    P_{ij} = P(\Phi_{j}|\Phi_{i}) = \frac{\exp(M_{ij})}{\sum_{j'}\exp(M_{ij'})}.
\end{align}
The round trip probability (Markov Chain) of walking from $i$ to $j$ can then be defined as:
\vspace{-0.1cm}
\begin{align}
    P_{ij}^{round} = \sum_{k \in \{1,\ldots,B\}}P_{ik}P_{kj}.
\end{align}
We further extend this into an unsupervised loss by encouraging the walker to walk back to its starting batch index $i$. This can be achieved by leveraging the fact that batch index implicitly refers to an object class $C_{obj} \in \{1,\ldots,O\}$ and $O \ll B$. An unsupervised loss $\loss_{asso}$ can be obtained as:
\vspace{-0.1cm}
\begin{align} 
    \loss_{asso}(\phi_{\theta}) = \left\|U-P^{round}\right\|_{F}^{2},
\end{align}
where $\|\cdot\|_F$ is the Frobenius norm and $U$ is a diagonal matrix of $\frac{1}{O}$ values: The $i$-th diagonal entry $U_{ii}$ represents that the walker starts at and returns to state $i$. 
$U$ can be adjusted if dataset is class-imbalanced.

  \begin{algorithm}[b]
  \caption{Collaborative learning algorithm}
  \label{alg:optim}
  \begin{algorithmic}[1]
    \Require{
      $\bx$ : input image,
      $P$ : network iteration
    }
    \Function{Optimise}{$\loss_{Total}$}
    \State{$\mathbf{r}_{hand} \leftarrow \text{ENC}_{hand}(\mathbf{x})$ } \Comment{Extract hand features}
    \State{$\mathbf{m}_{hand} \leftarrow $g$\textsuperscript{\textit{HME}}(\mathbf{r}_{hand})$} \Comment{Get hand mesh}
    \For{$t=1$ to $P$}
        \State{$\phi_{hand} \leftarrow g_{hand}^{conv}(\mathbf{m}_{hand})$}  \Comment{Hand Graph Conv.}
        \State{$\mathbf{r}_{obj} \leftarrow \text{ENC}_{obj}(\mathbf{x}) + \phi_{hand}$ } \Comment{Feature update}
        \State{$\mathbf{m}_{obj} \leftarrow $g$\textsuperscript{\textit{OME}}(\mathbf{r}_{obj})$} \Comment{Get object mesh}
        \State{$\phi_{obj} \leftarrow g_{obj}^{conv}(\mathbf{m}_{obj})$} \Comment{Object Graph Conv.}
        \State{$\mathbf{r}_{hand}{'} \leftarrow \mathbf{r}_{hand} + \phi_{obj}$} \Comment{Feature update}
        \State{$\mathbf{m}_{hand} \leftarrow $g$\textsuperscript{\textit{HME}}(\mathbf{r}_{hand}{'})$} 
    \EndFor
    \EndFunction
  \end{algorithmic}
\end{algorithm}


\section{Experiments}
\vspace{-0.1cm}
\paragraph{Implementation details.}
We implement our method in PyTorch~\cite{PyTorch}.
All experiments are run on an Intel i9-CPU @ 3.50GHZ, 16 GB RAM, and one NVIDIA RTX 3090 GPU. 
We train all parts of the network simultaneously with Adam optimiser \cite{kingma2014adam} at a learning rate $10^{-4}$ for 400 epochs. We then freeze the ResNet \cite{he2016deep} encoders and decrease the learning rate to $10^{-5}$ for another 100 epochs. We empirically fixed $K=3$ attention heads and $P=2$ iterations to produce the best results. Our final loss $\loss_{final}$ is defined as: 
\vspace{-0.1cm}
\begin{align}
    \loss_{final} = \loss_{hand} + \loss_{obj} + \loss_{asso}.
\end{align}

\vspace{-0.1cm}
\paragraph{Datasets.}
\emph{First-person hand benchmark (FHB).} This is a widely-used dataset \cite{FirstPersonAction_CVPR2018} which contains egocentric RGB-D videos on a wide range of hand-object interactions. The ground-truth of hand and object poses are captured via magnetic sensors. There are $4$ available objects, \ie juice bottle, liquid soap, milk and salt. For fair comparisons with \cite{tekin2019h+,hasson2020leveraging}, we follow the same \textit{action split} for evaluation where each object is present in both training and testing. We also compare with \cite{hasson2019learning} which uses the \textit{subject split} of the dataset following their experimental settings: They filtered frames when the hand is further than $1cm$ away from the manipulated object and excluded the milk object. We call this subset \emph{FHB$^{-}$} which contains a total of $3$ objects. 

\emph{ObMan.} This is a large synthetic dataset \cite{hasson2019learning} which was produced by rendering hand meshes with selected objects from ShapeNet \cite{chang2015shapenet}. 
It captures 8 object categories and results in a total of 2,772 meshes which are split among 154,000 image frames. We pretrained the network on \emph{ObMan} before training on other real datasets: We observed in our preliminary experiments that their setting led to consistent improvements over training directly on real data.

\emph{DexYCB.} This is a recent real dataset for capturing hand grasping of objects \cite{chao2021dexycb}. It consists a total of 582,000 image frames on 20 objects from YCB-Video dataset \cite{xiang2017posecnn}. We present results on all $4$ official dataset split settings. 

\emph{HO-3D.} \cite{hampali2020honnotate} is most similar to \emph{DexYCB} where it consists of 78,000 images frames on 10 objects. We present results on the official dataset split (version 2). The hand mesh error is reported after procrustes alignment and in $mm$.

\paragraph{Evaluation metrics.}
\emph{Hand error.} We report the mean end-point error ($mm$) over 21 joints and use the percentage of correct keypoints (PCK) score to evaluate at different error thresholds.

\emph{Object error.} We measure the accuracy of object reconstruction by computing the Chamfer distance ($mm$) between points sampled on ground truth and predicted mesh.

\emph{Hand-object interaction.} To understand hand-object interaction, we followed \cite{hasson2019learning} to include penetration depth $(mm)$ and intersection volume $(cm^3)$. Penetration depth refers to the maximum distances from hand mesh vertices to the object’s surface when in a collision. Intersection volume is obtained by voxelising the hand and object using a voxel size of $0.5cm$.


\paragraph{Results.}
\begin{figure}[t] 
\centering
\includegraphics[width=0.85\linewidth]{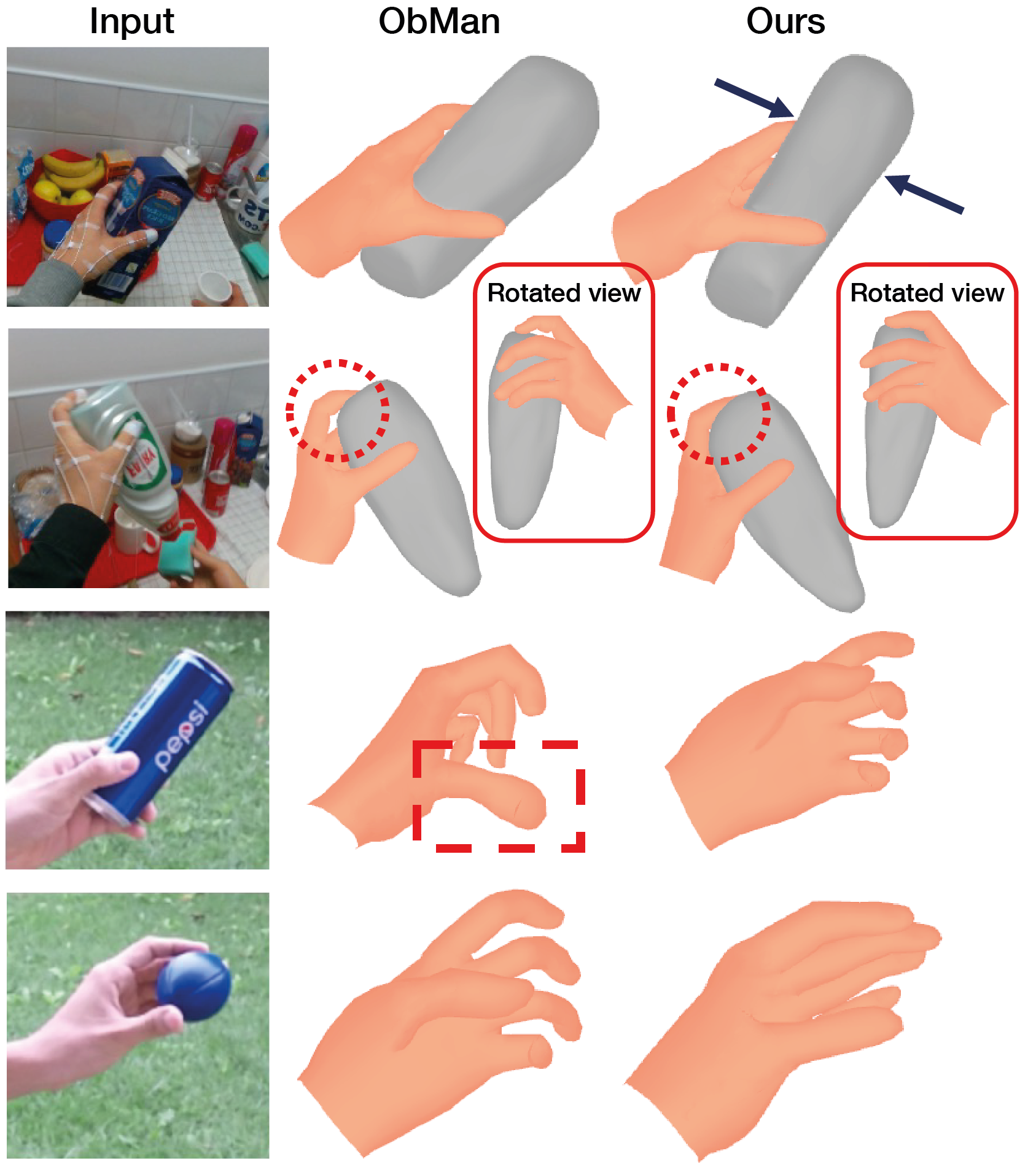}
\vspace{-0.3cm}
\caption{Qualitative comparison with ObMan \cite{hasson2019learning}. Top two rows refers to models trained with \emph{FHB}. Bottom two rows refers to in-the-wild settings where models are only trained with synthetic dataset \emph{ObMan}. Our method is able to refine and sharpen object mesh under the collaborative learning framework (see blue arrows) and generalise better hand pose in both settings.
}
\label{fig:qualitative}
\end{figure}
\emph{Joint hand-object reconstruction.} 
As recent efforts on joint hand-object reconstructions \cite{hasson2020leveraging,cao2020reconstructing,hasson20_handobjectconsist,karunratanakul2020grasping,yang2021cpf} assume known object models, we compare with \cite{hasson2019learning} (adopted differential MANO model, AtlasNet and does not assume known object models) in Table \ref{table:comparison to ObMan}. Similar to \emph{FHB}, we used the default \emph{DexYCB} split and filtered frames when hand and manipulated object are $1cm$ apart. We name this subset to be \emph{DexYCB$^{-}$} and retrain \cite{hasson2019learning} using their released code. As shown, there is still a presence of interpenetration at test time and even increases the hand error by $0.7mm$ on \emph{FHB$^{-}$} with contact loss in \cite{hasson2019learning}. This is mainly due to the fact that their model is not implicitly learning the physical rules imposed by the contact loss. In contrast, our method consistently outperforms \cite{hasson2019learning} with a higher hand-object reconstruction accuracy. In addition, we provide qualitative comparisons on \emph{FHB} and \emph{CORe50} \cite{lomonaco2017core50} datasets in Fig. \ref{fig:qualitative}.
\begin{table}[]
\begin{center}
\caption{Quantitative comparison with ObMan \cite{hasson2019learning} on \emph{ObMan}, \emph{FHB$^{-}$} and \emph{DexYCB$^{-}$} datasets. $^{*}$ refers to the results with contact loss. Our proposed collaborative learning strategy performs competitively without physical contact loss.}
\vspace{-0.3cm}
\label{table:comparison to ObMan}
\resizebox{1.02\linewidth}{!}{
\begin{tabular}{l | ccc | ccc |cc}\toprule
Datasets      & \multicolumn{3}{c|}{\emph{ObMan}}                                     & \multicolumn{3}{c|}{\emph{FHB$^{-}$}}  &
      \multicolumn{2}{c}{\emph{DexYCB$^{-}$}} \\
Method & \cite{hasson2019learning} & \cite{hasson2019learning}* & Ours & \cite{hasson2019learning} & \cite{hasson2019learning}* & Ours & \cite{hasson2019learning}* & Ours \\ \hline
Hand error ($mm$)$\downarrow$ & 11.6 & 11.6 & \textbf{9.1} & 28.1 & 28.8 & \textbf{25.3} & 17.6 & \textbf{15.3} \\ 
Object error ($mm$)$\downarrow$ & 641.5 & 637.9 & \textbf{385.7} & 1579.2 & 1565.0 & \textbf{1445.0} & 549.4 & \textbf{501.2}\\
Max. penetration ($mm$)$\downarrow$ & 9.5 & 9.2 & \textbf{7.4} & 18.7 & \textbf{12.1} & 16.1 & 14.6 & \textbf{12.1}\\ 
Intersection vol. ($cm^{3}$)$\downarrow$ & 12.3 & 12.2 & \textbf{9.3} & 26.9 & 16.1 & \textbf{14.7} & 14.9 & \textbf{13.4}\\ \bottomrule
\end{tabular}}
\end{center}
\end{table}
\newcolumntype{C}{>{\centering\arraybackslash}X}
\begin{table}
\begin{center}
\vspace{-0.3cm}
\caption{
Error rates of different hand pose estimation methods on \emph{HO-3D}. Note that the reported results for \cite{liu2021semi} output hand meshes only. We outperform two other architecturally similar networks \cite{hasson2019learning,hasson2020leveraging} without known object models under our collaborative learning framework.}
\label{table:HO3D}
\vspace{-0.2cm}
\resizebox{0.9\linewidth}{!}{%
\begin{tabularx}{\linewidth}{r | C  C  C  C}
\toprule
        & Mesh & F-score & F-score & Known \\ 
 Method & error $\downarrow$ & @$5mm \uparrow$  & @$15mm \uparrow$ & objects \\
\midrule 
\cite{hasson2019learning} & 11.0 & 46.0 & 93.0 & \xmark \\
\cite{hampali2020honnotate} & 10.6 & 50.6 & 94.2 & \cmark \\
\cite{liu2021semi} & \textbf{9.5} & \textbf{52.6} & \textbf{95.5} & \cmark \\
\cite{hasson2020leveraging} & 11.4 & 42.5 & 93.4 & \cmark \\
\midrule
Ours  & 10.9 & 48.5 & 94.3 & \xmark \\
\bottomrule
\end{tabularx}%
}
\end{center}
\vspace{-0.6cm}
\end{table}

\begin{figure}
\centering
\includegraphics[width=1\linewidth]{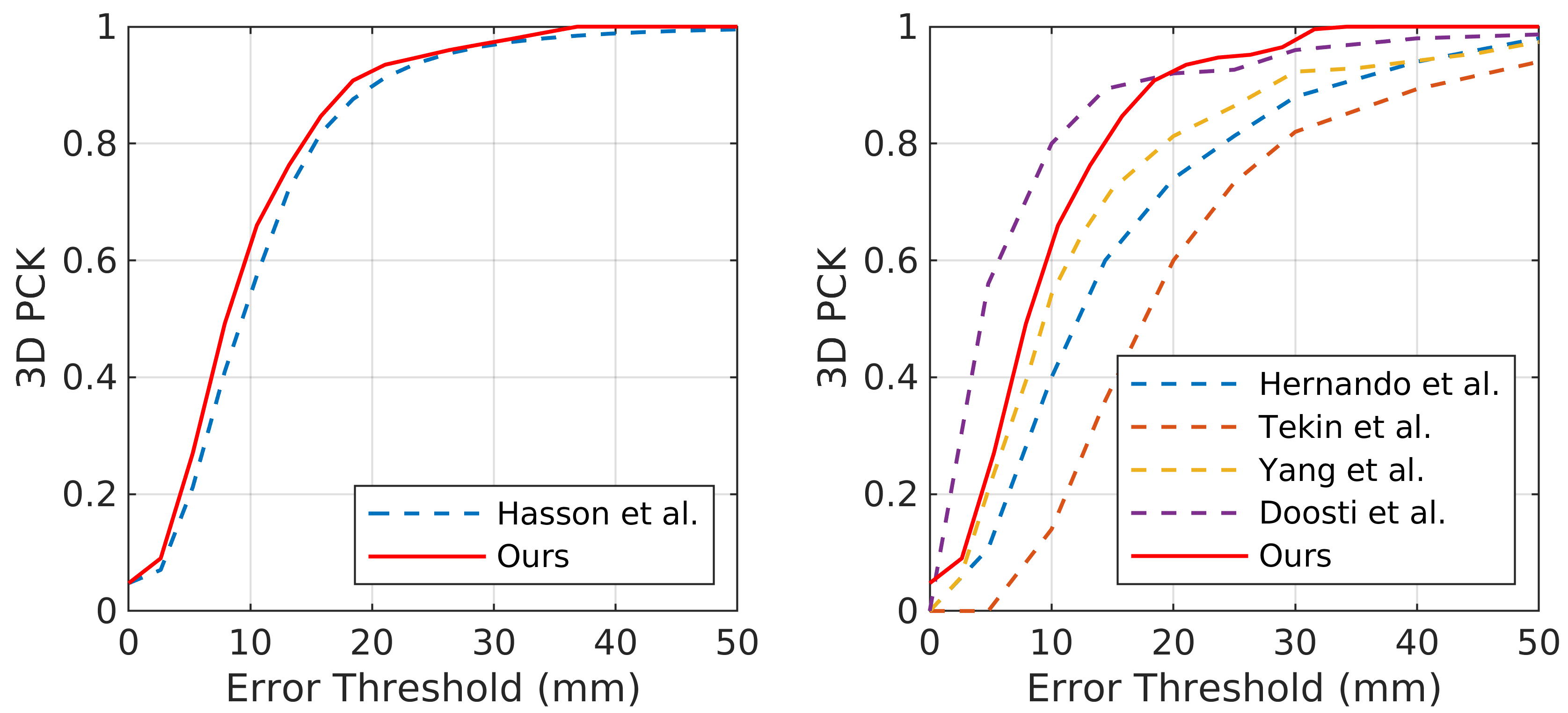}
\vspace{-0.6cm}
\caption{3D PCK for \emph{ObMan} (\textit{left}) and \emph{FHB} (\textit{right}). Note that Hasson \etal refers to \cite{hasson2019learning}, and Doosti \etal~\cite{doosti2020hope} is a hand-object pose estimation method where known object is given.
\vspace{-0.3cm}
}
\label{fig:pck}
\end{figure}

\emph{Hand pose estimation.} We first compare with state-of-the-art methods on \emph{HO-3D} \cite{hampali2020honnotate} in Table \ref{table:HO3D}. As shown, our method performs competitively against methods that assumes known object models. Then, we compare on \emph{FHB} (both \textit{action split} and \textit{subject split}) in Table \ref{table:2 FHB} and \ref{table:PCK on FHB}. Note that \cite{hasson2020leveraging} is an extension to \cite{hasson2019learning} which leverages photometric consistency but required known object model. As shown in Table \ref{table:2 FHB}, we demonstrate superior performances among all three architecturally similar networks \cite{hasson2019learning,hasson2020leveraging}. We attribute the performance gain in \textit{action split} (\ie \emph{FHB}) to the fact that \emph{FHB$^{-}$} contains almost half of \emph{FHB} with incomplete object list and unseen test subjects during test time. We analyse our hand pose estimation performance using the PCK metric in Table \ref{table:PCK on FHB}. Note that Yang \etal~\cite{yang2020collaborative} takes sequential images as input and leverages action recognition task in their collaborative framework. We achieve state-of-the-art performance to in hand pose estimation with the advantage of object reconstruction. 3D PCK curves are shown in \fig{pck}. Finally, we compare with a supervised version of Spurr \etal~\cite{spurr2020} which won the HANDS 2019 Challenge \cite{armagan2020measuring} on \emph{DexYCB} \cite{chao2021dexycb}. In Table \ref{table:DexYCB}, the numbers are obtained from \cite{chao2021dexycb} where \cite{spurr2020} has a HRNet32 \cite{sun2019deep} backbone.

\newcolumntype{C}{>{\centering\arraybackslash}X}
\begin{table}
\begin{center}
\caption{Error rates of different algorithms. \emph{FHB} refers to \textit{action split} and \emph{FHB$^{-}$} refers to \textit{subject split} of the dataset.}
\label{table:2 FHB}
\vspace{-0.2cm}
\resizebox{0.9\linewidth}{!}{%
\begin{tabularx}{\linewidth}{r | C | C}
\toprule
       & \emph{FHB}        & \emph{FHB$^{-}$} \\ 
Method & Hand Error & Hand Error \\ 
\midrule 
Tekin \etal~\cite{tekin2019h+}  & 15.8       & -   \\
Hasson \etal~\cite{hasson2020leveraging} & -       & 28.0 \\
Hasson \etal~\cite{hasson2019learning}& 18.0         & 27.4\\
Cao \etal~\cite{cao2020reconstructing}& 14.2         & -\\
\midrule
Ours  & \textbf{9.8}        & \textbf{25.3}  \\
\bottomrule
\end{tabularx}%
}
\end{center}
\vspace{-0.2cm}
\end{table}
\newcolumntype{C}{>{\centering\arraybackslash}X}
\begin{table}
\begin{center}
\vspace{-0.2cm}
\caption{PCK performance over respective error threshold on \emph{FHB}. Compared to another collaborative learning framework \cite{yang2020collaborative} and graph-based method \cite{doosti2020hope}, our method performs better and is able to reconstruct both hand-object meshes.}
\label{table:PCK on FHB}
\vspace{-0.2cm}
\resizebox{0.9\linewidth}{!}{%
\begin{tabularx}{\linewidth}{r | C | C}
\toprule
Method & PCK@20mm  & PCK@25mm\\ 
\midrule 
Tekin \etal~\cite{tekin2019h+}  & 69.17\% & 81.25\% \\
Hernando \etal~\cite{FirstPersonAction_CVPR2018}  & 74.73 \% & 82.10\% \\
Yang \etal~\cite{yang2020collaborative}  & 81.03\% & 86.61\%\\
Doosti \etal~\cite{doosti2020hope} & 92.17\% & 92.63\%\\
\midrule
Ours  & \textbf{93.14\%} & \textbf{95.65\%}\\
\bottomrule
\end{tabularx}%
\vspace{-0.2cm}
}
\end{center}
\end{table}
\vspace{-0.1cm}
\newcolumntype{C}{>{\centering\arraybackslash}X}
\begin{table}
\begin{center}
\caption{Error rates on \emph{DexYCB} and \cite{spurr2020} is the winner of HANDS 2019 Challenge \cite{armagan2020measuring}. 
Table indicates hand error ($mm$) with AUC values in parentheses.
S0-S3 are the official dataset splits \cite{chao2021dexycb}. 
}
\label{table:DexYCB}
\vspace{-0.2cm}
\resizebox{1\linewidth}{!}{%
\begin{tabularx}{\linewidth}{l | C | C | C | C}
\toprule
 & S0 & S1 & S2 & S3 \\ 
\midrule 
\cite{spurr2020} & {\footnotesize17.34(0.698)} & {\footnotesize22.26(0.615)} & \textbf{{\footnotesize25.49(0.530)}} & {\footnotesize18.44(0.686)} \\
Ours  & \textbf{{\footnotesize16.05(0.722)}} & \textbf{{\footnotesize21.22(0.620)}} & {\footnotesize27.01(0.521)} & {\footnotesize\textbf{17.93(0.698)}}\\
\bottomrule
\end{tabularx}
}
\end{center}
\vspace{-0.4cm}
\end{table}

\paragraph{Ablation study.}
To motivate our design choices, we present a quantitative comparison of our method with various components disabled. We validate that the combination of our design choices outperforms the na\"ive collaborative learning baseline (see supplementary), which predicts the embeddings directly and perform 3D reconstruction last. 

\begin{table}[]
\begin{center}
\vspace{-0.1cm}
\caption{Performances of different network design choices on \emph{FHB$^{-}$}. We experiment on network iterations $P$, associative loss $\loss_{asso}$ and different convolution operators. The baseline on the first row is same as ObMan \cite{hasson2019learning}.}
\label{table:ablation_P}
\vspace{-0.25cm}
\resizebox{1\linewidth}{!}{
\begin{tabular}{l | c  c | c  c} \hline
       & \multicolumn{2}{c|}{w $\loss_{asso}$}        & \multicolumn{2}{c}{w/o $\loss_{asso}$} \\ 
Method & Hand Error & Object Error & Hand Error & Object Error\\ \hline
Baseline & - & - & 28.4 & 1655.2       \\
Baseline ($P=1$) & 26.9 & 1600.3 & 27.4 & 1625.9 \\
Baseline ($P=2$) & \textbf{25.3} & \textbf{1445.0}  & 26.3 & 1618.4 \\ 
Baseline ($P=3$) & 25.4 & 1448.2 & 26.4 & 1620.5 \\ 
Baseline ($P=4$) & 25.3 & 1447.9 & 26.3 & 1612.9 \\ 
Baseline ($P=5$) & 25.3 & 1445.6 & 26.2 & 1618.8 \\ \hline
GCN \cite{kipf2016semi} ($P=1$)& 27.1 & 1587.6 & 27.8 & 1629.8       \\
GCN \cite{kipf2016semi} ($P=2$)& 27.0 & 1590.8 & 28.2 & 1635.1       \\
Spiral \cite{gong2019spiralnet++,kulon2020weakly} ($P=1$) & 26.8 & 1581.8 & 27.6 & 1630.1 \\
Spiral \cite{gong2019spiralnet++,kulon2020weakly} ($P=2$) & 26.9 & 1600.2 & 27.6 & 1629.5\\ \hline
\end{tabular}}
\end{center}
\vspace{-0.6cm}
\end{table}

\emph{Impact of the number of network iterations ($P$):} Table \ref{table:ablation_P} shows the results of varying $P$ with associative loss and demonstrate that associative loss contributes to improving hand and object error. 
This can be expected since hand-object reconstruction are highly correlated such that learning
in a collaborative manner enables performance boost to each other. The effectiveness of our proposed dynamic graph convolution can be demonstrated by the fast performance saturation at $P=2$. Note that we took \cite{hasson2019learning} as our baseline and graph convolution is enabled from $P=1$. 

\emph{Comparison with static graph convolution:} To motivate our proposed dynamic graph convolution, we experiment with two commonly used graph convolution in Table \ref{table:ablation_P}, \ie GCN \cite{kipf2016semi} and spiral mesh convolution \cite{gong2019spiralnet++,kulon2020weakly}. As the graph convolutions weights are only updated after $P$ iterations, increasing network iterations will have zero effects. It can be seen that static graph convolution does not benefit from increasing network iterations. We also observed that our unsupervised associative loss ($\loss_{asso}$) consistently improves hand-object error across Table \ref{table:ablation_P}.

\begin{figure}
\centering
\includegraphics[width=1\linewidth]{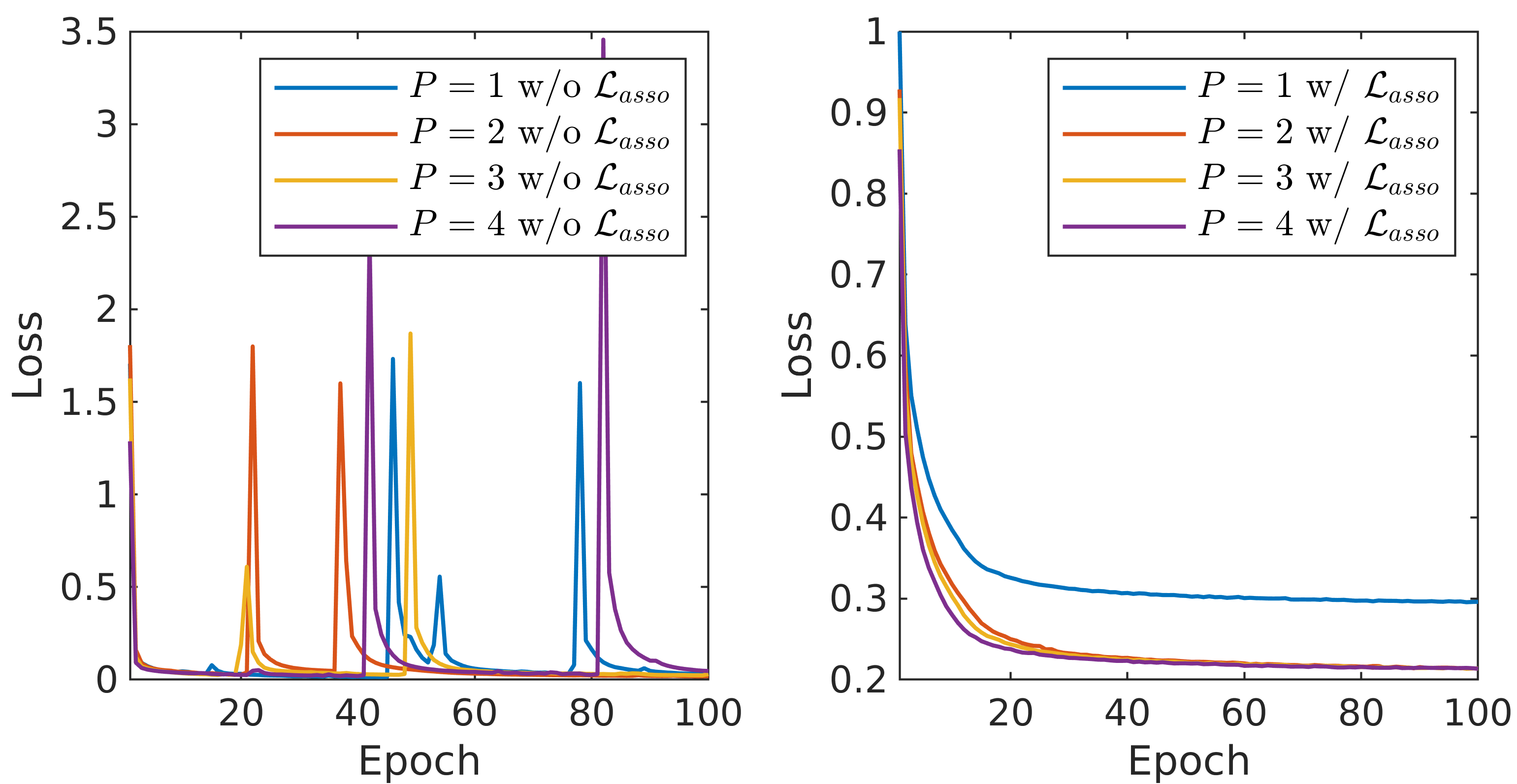}
\vspace{-0.6cm}
\caption{Progression of training losses for iterations $P=\{1,\dots,4\}$, without (\textit{left}) and with (\textit{right}) associative loss $\loss_{asso}$.
}
\vspace{-0.2cm}
\label{fig:converge}
\end{figure}

\emph{Effectiveness of associative loss ($\loss_{asso}$):} To further study the effect of our unsupervised $\loss_{asso}$, we plot the training loss for the collaborative framework, with and without associative loss in \fig{converge}. Unsurprisingly, we find that increasing network iterations $P$ contributes to a higher convergence rate (right of \fig{converge}). We also observe that our unsupervised associative loss ($\loss_{asso}$) is able to stabilise the training across all iterations (left of \fig{converge}). This shows that training with $\loss_{asso}$ is crucial for this framework.

\begin{table}[]
\centering
\caption{Ablation studies on collaborative learning framework design. We experiment on both \emph{FHB$^{-}$} and the default \emph{DexYCB} (S0) dataset split. * refers to the na\"ive collaborative learning baseline.}
\vspace{-0.2cm}
\label{table:ablation_CL}
\resizebox{1\linewidth}{!}{
\begin{tabular}{ll | c  c | c  c}\toprule
            &              & \multicolumn{2}{c|}{\emph{FHB$^{-}$}}   & \multicolumn{2}{c}{\emph{DexYCB} (S0)} \\
\multicolumn{2}{l|}{Method} & Hand Error & Object Error & Hand Error  & Object Error \\ \hline
$P=1$ & Ours* & 28.0 & 1759.4 & 17.9 & 563.4 \\
      & Ours  & \textbf{26.9} & \textbf{1600.3} & \textbf{17.6} & \textbf{529.3} \\ \hline
$P=2$ & Ours* & 27.6 & 1726.8 & 17.5 & 554.6 \\
      & Ours  & \textbf{25.3} & \textbf{1445.0} & \textbf{16.1} & \textbf{461.1} \\ \hline
$P=3$ & Ours* & 27.1 & 1678.1 & 17.3 & 542.1 \\
      & Ours  & \textbf{25.4} & \textbf{1448.2} & \textbf{16.0} & \textbf{464.2} \\  \bottomrule  
\vspace{-1cm}
\end{tabular}}
\end{table}

\emph{Mesh generation within iterations:} We target the problem of \textbf{mutual occlusion} of interacting hand and object by sharing 3D information at each iteration via graph convolution. To validate this design choice, we construct a simpler collaborative learning framework which directly predicts embeddings $\phi_{\theta}$ and reconstruct meshes $m_{\theta}$ at the final stage (see supplementary diagram). As \emph{FHB} has limited backgrounds and visible magnetic sensors, we compare the
two design on \emph{FHB} and \emph{DexYCB}. Table \ref{table:ablation_CL} shows that 
our final design consistently outperforms the na\"ive composition baseline across both datasets. We observe that sharing 3D mesh information across hand and object branches improves both reconstruction performance. 
At the bottom right of \fig{framework}, we provide a qualitative example of how reconstruction changes with graph convolution. It can be confirmed that our attention-guided
graph convolution combined with collaborative learning enables better mesh quality as well as more accurate pose estimation. We provide additional qualitative results in \fig{qualitative_end}.

\noindent
\begin{minipage}{1\textwidth}
  \strut\newline
  \centering
  \includegraphics[width=1\textwidth]{{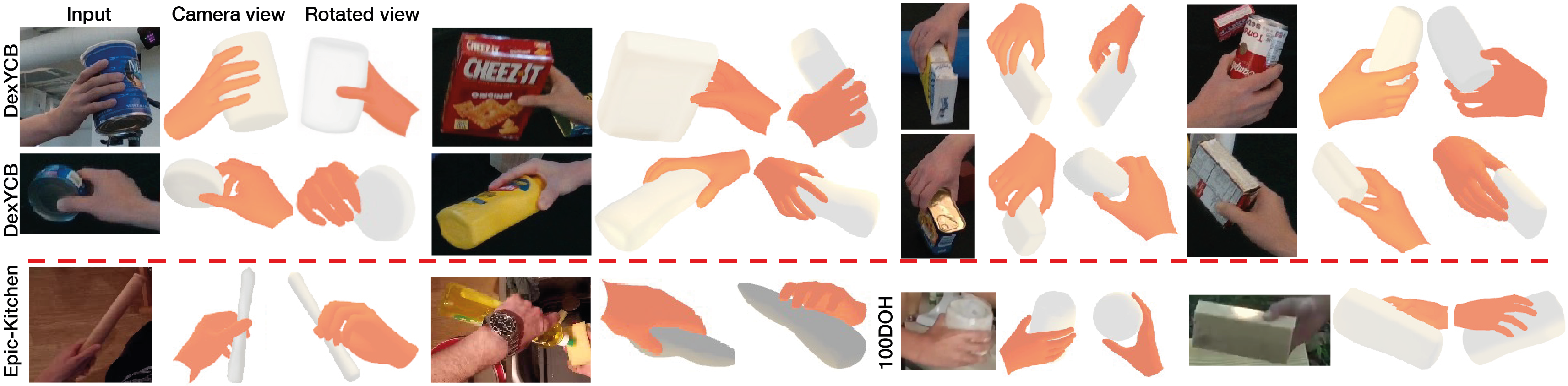}}
  \vspace{-0.6cm}
  \captionof{figure}{Qualitative results on \emph{DexYCB} (top two rows), \emph{EPIC-Kitchens} \cite{Damen2018EPICKITCHENS} (left of bottom row) and \emph{100 Days of Hands} (100DOH) \cite{Shan20} (right of bottom row). The bottom row refers to in-the-wild settings. Our model, trained only on \emph{DexYCB}, shows robustness to various hand poses, objects and scenes.}\label{fig:qualitative_end}
\end{minipage}


\vspace{-0.75cm}
\section{Conclusion}
In this paper, we have proposed a novel collaborative learning framework which allows the sharing of mesh information across hand and object branches iteratively. The main idea behind this study was to demonstrate that mutual occlusion can be tackled in a learning-based strategy.
We designed an attention-guided graph convolution which captures long-range dependencies from dynamic graph in a single layer. 
However, training with increasing network iterations can be highly unstable. Therefore, we proposed an unsupervised associative loss to stabilise the training and improve the feature transferring process.
Our method demonstrated superior performance when compared to other existing approaches on multiple widely-used datasets.

\emph{Limitations.} 
Our work relied on AtlasNet for object reconstruction, and we observed that the object reconstruction quality varies with the size of training data. 
Furthermore, we have only considered static objects, hence future works should consider the interaction between hands and articulated objects.

\emph{Potential negative societal impact.} 
Our method can facilitate hand-based interaction in various applications including augmented and virtual reality. In general, advances in hand-based interaction can potentially introduce a barrier to or discourage people having difficulty in using their hands. This could be mitigated when accompanied by technical advances in other modes of interaction, \eg eye or mouse tracking, or body gesture-based interaction.

\vspace{-0.1cm}
\section*{Acknowledgements}
\vspace{-0.2cm}
\footnotesize \noindent This research was supported by the Ministry of Science and ICT, Korea, under the Information Technology Research Center (ITRC) support program (IITP-2022-2020-0-01789) supervised by the Institute for Information \& Communications Technology Planning \& Evaluation (IITP) and an IITP grant (2021-0-00537). The computations described in this research were performed using the Baskerville Tier 2 HPC service (https://www.baskerville.ac.uk/) that was funded by EPSRC Grant EP/T022221/1 and is operated by Advanced Research Computing at the University of Birmingham. KIK was supported by the National Research Foundation of Korea (NRF) grant (No. 2021R1A2C2012195) funded by the Korea government (MSIT).

\clearpage
{\small
\bibliographystyle{ieee_fullname}
\bibliography{iccv_elden}
}

\end{document}